# Multi-fidelity surrogate with heterogeneous input spaces for modeling melt pools in laser-directed energy deposition

Nandana Menon[1] and Amrita Basak[1*]

[1]Department of Mechanical Engineering, The Pennsylvania State University, University Park, 16802, PA, USA.

*Corresponding author: E-mail: aub1526@psu.edu; Contributing author: nfm5316@psu.edu;

## Abstract

Multi-fidelity (MF) modeling is a powerful statistical approach that can intelligently blend data from varied fidelity sources. This approach finds a compelling application in predicting melt pool geometry for laser-directed energy deposition (L-DED). One major challenge in using MF surrogates to merge a hierarchy of melt pool models is the variability in input spaces. To address this challenge, this paper introduces a novel approach for constructing an MF surrogate for predicting melt pool geometry by integrating models of varying complexity, that operate on heterogeneous input spaces. The first thermal model incorporates five input parameters i.e., laser power, scan velocity, powder flow rate, carrier gas flow rate, and nozzle height. In contrast, the second thermal model can only handle laser power and scan velocity. A mapping is established between the heterogeneous input spaces so that the five-dimensional space can be morphed into a pseudo two-dimensional space. Predictions are then blended using a Gaussian process-based co-kriging method. The resulting heterogeneous multi-fidelity Gaussian process (Het-MFGP) surrogate not only improves predictive accuracy but also offers computational efficiency by reducing evaluations required from the high-dimensional, high-fidelity thermal model. The tested Het-MFGP yields an $R^2$ of 0.975 for predicting melt pool depth. This surpasses the comparatively modest $R^2$ of 0.592 achieved by a GP trained exclusively on high-dimensional, high-fidelity data. Similarly, in the prediction of melt pool width, the Het-MFGP excels with an $R^2$ of 0.943, outshining the GP's performance, which registers a lower $R^2$ of 0.588. The results underscore the benefits of employing Het-MFGP for modeling melt pool behavior in L-DED. The framework successfully demonstrates how to leverage multimodal data and handle scenarios where certain input parameters may be difficult to model or measure.

## 1. Introduction

The characterization of melt pools in laser-directed energy deposition (L-DED) metal additive manufacturing (AM) is a crucial requirement for controlling the build properties. The melt pool shape is reported to affect the local solidification rates as well as the grain structures that form in the fusion zone [1]. However, understanding melt pool response to the varied process parameters that govern the deposition process traditionally requires many experiments. Conducting physical experiments and extracting the data can be tedious and repetitive, requiring multiple iterations as the material or machine changes. As an alternative, physics-based models have been extensively developed and explored for faster generation of melt pool data. These models range in complexity and accuracy – from analytical and numerical models that provide fast, low-resolution solutions to complex computational fluid dynamics models that solve coupled fluid flow and heat transfer during the process [2]. Moreover, the success of machine learning (ML) in recent times has percolated into the field of L-DED as well to develop digital twins that can speed up melt pool predictions. Multi-fidelity (MF) surrogate presents itself as a promising ML technique that can be built by blending information from hierarchical models [3]. This method balances accuracy and computational efficiency, and therefore, is particularly useful for modeling in L-DED, where the design space can be large, varying with machine and material.

MF surrogates can be deterministic or non-deterministic based on the statistical approach employed to evaluate the surrogate parameters. Typically, deterministic methods evaluate the parameters via calibration to minimize the discrepancy between the model fidelities [4]. Non-deterministic surrogates employ probabilistic models such as Gaussian process (GP). GPs are a class of stochastic processes identified by a mean and a covariance [5]. The surrogate thus developed is also characterized by a mean, which is the most probable representation of the data, and a covariance that quantifies the uncertainty associated with the predictions. Such GP surrogates can be extended into an MF scheme using techniques, such as co-kriging, which enable the integration of models of multiple fidelities by using a covariance function to couple them within a GP framework [6]. By combining MF data, MFGPs can provide more accurate predictions compared to using any one type of data in isolation. Additionally, MFGPs provide a principled framework for quantifying uncertainty – they not only provide predictions of the underlying function but also estimate the associated confidence in the prediction.

MFGPs have been previously implemented for both laser-powder bed fusion (L-PBF) and L-DED. Menon et al. [7] used MFGPs to predict melt pool geometry in L-DED. The MF surrogate was able to achieve a 55% improvement in accuracy compared to a single-fidelity GP trained on the same quantity of high-fidelity data. An extension of the work involved integrating experimental data with numerical melt pool data to generate predictions within 5% of the actual values [8]. Saunders et al. [9] attempted to use nonlinear autoregressive MFGP as a regressor and classifier. While the regressor made reasonable predictions of melt pool geometry, the classifier was not equally competent at determining printability. Saunders et al. [10] also explored MFGPs using multi-output GPs for the creation of process-structure-property linkages for L-PBF of SS316L. The study uses a space-filling Latin Hypercube Sampling (LHS) sequential design for each level of fidelity, starting with experiments, finite-element, numerical, and analytical models, in decreasing order of fidelity.

Classical MFGPs, such as those discussed, consist of models belonging to the same input space and rarely consider the situation where the fidelities belong to heterogeneous input spaces. In practice, MF data can result from altering input space definitions through (i) changes in dimensionality or (ii) input spaces originating from diverse distributions with distinct features. Changes in dimensionality can arise when a model and its reduced order representation are considered. Another situation is when the input space is simplified for a low-fidelity representation by either down selecting a few parameters or by combining inputs to form new dimensionless variables. Similarly, when dealing with multimodal data for the same output, e.g., temperature data from cameras and thermocouples, heterogeneity is introduced as the two datasets are defined by different input features. Such diverse spaces cannot be directly combined into MF frameworks. One simple technique explored in literature is the construction of a homogeneous representation of the heterogeneous domains by transforming the input spaces such that the distance between the outputs is minimized. Tao et al. [4] introduced the Input Mapping Calibration (IMC) method where the high-dimensional input space was adapted to the low-dimensional input space by minimizing the distance between their respective outputs. Wang et al. [11] introduced heterogeneous space adaptation for a classification problem using manifold alignment. Their work focused on constructing a common latent space for all inputs and using the labels to understand the alignment. More recently, Liu et al. [12] used a multi-task GP for knowledge transfer across correlated tasks by using Bayesian calibration to achieve input alignment across heterogeneous spaces. Sarkar et al. [13] used MFGP across heterogeneous input spaces by learning an asymmetric mapping using a multi-input multi-output regressor on the data with corresponding heterogeneous input pairs across fidelities. Hebbal et al. [14] employed deep MFGPs based on the work of Cutajar et al. [15], incorporating the mapping between the input spaces of different fidelity in a non-parametric way. However, deep GPs and, hence, their multi-fidelity counterparts have the bottleneck of having to determine the layers, and the proportion of data from each fidelity in a mini batch amongst other hyperparameters pertaining to deep learning models.

For predicting melt pool properties in L-DED, low-dimensional, low-fidelity (LF) models can be developed by neglecting or consolidating specific parameters to create simplified representations of the underlying physical phenomena [16]. Consequently, the dimensions of the input space of a high-dimensional, high-fidelity (HF) model, which encompasses a broader range of process parameters, is greater than its LF counterpart. Another scenario contributing to heterogeneity arises when certain printing parameters are either unknown or challenging to measure within a design of experiments (DOE). While the current study focuses on the first scenario, it can also be extended to the practical challenge observed in the latter. For this, the paper introduces the development of a heterogeneous MFGP (Het-MFGP) surrogate, constructed utilizing melt pool models that operate within diverse input spaces, to predict melt pool geometry. The schematic representation of the Het-MFGP framework is depicted in Fig. 1. The framework starts by developing an HF thermal model validated using experimental data. The LF model is calibrated to mimic the HF model, as discussed in Section 2.1. The input spaces of these two models are heterogeneous as they belong to spaces of different dimensions, defined in $\mathbb{R}^{d_{HF}}$ and $\mathbb{R}^{d_{LF}}$, respectively. Since $d_{HF} > d_{LF}$, a mapping between the heterogeneous input spaces, based on IMC [4], is developed to transform the high-dimensional input space into a pseudo space of dimension $d_{LF}$. This process involves optimizing the loss function, as discussed in Section 2.3, such that the difference between the outputs produced by the LF thermal model for pseudo low-dimensional space, and HF thermal model for the original high-dimensional input space is minimized. Thereafter, the Het-MFGP surrogate is calibrated (Sections 3.1 and 3.2) and compared against single-fidelity counterparts to establish its performance (Section 3.3). The results show that utilizing the Het-MFGP surrogate leads to significantly higher accuracy and confidence in contrast to a GP trained solely on HF data points, highlighting the advantageous performance of the Het-MFGP in modeling melt pool behavior in L-DED.

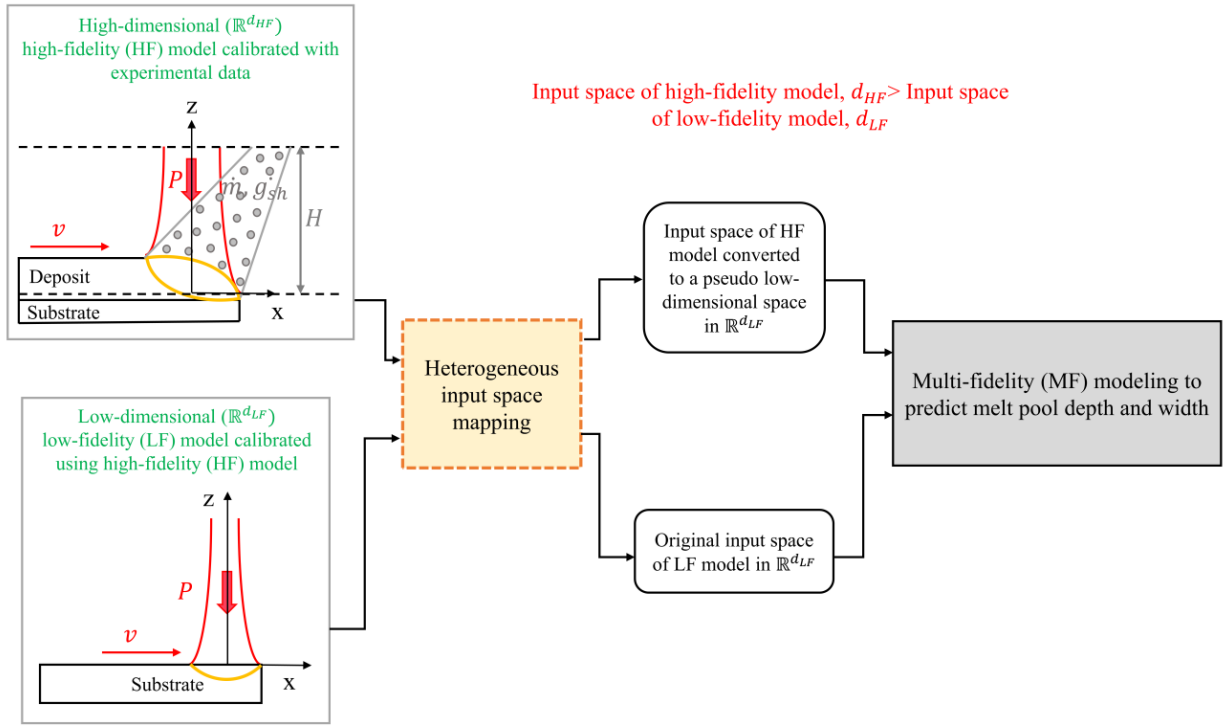

**Fig 1. The framework of the Het-MFGP. Two thermal models are selected to represent the high-dimensional high-fidelity (HF) model and the low-dimensional low-fidelity (LF) model. The HF model is validated using experimental data. The model parameters of the LF model are calibrated to closely mimic the results of the HF model. The HF model incorporates five process parameters – laser power ($P$), scan velocity ($v$), powder flow rate ($\dot{m}$), carrier gas flow rate ($\dot{g}_{sh}$), and nozzle height ($H$) while the LF model only takes in $P$ and $v$ as variable inputs. The HF and LF input spaces are heterogeneous in that they belong to spaces of different dimensions defined by $\mathbb{R}^{d_{HF}}$ and $\mathbb{R}^{d_{LF}}$, respectively. Since $d_{HF} > d_{LF}$, the HF space needs to be modified using a mapping between the heterogeneous spaces. The mapping converts the**

HF space into a pseudo space in $\mathbb{R}^{d_{LF}}$. After mapping, the dimensions of the input spaces of the thermal models are aligned and, therefore, ready for MF modeling.

## 2. Methodology
### 2.1 Thermal models for Het-MFGP
#### 2.1.1 High-dimensional high-fidelity (HF) thermal model

The HF model of the L-DED process is carried out using a formulation developed by Huang et al. [17] that presents a comprehensive analytical solution to the thermal distribution experienced during the powder-fed L-DED process. The analytical model couples a Gaussian energy distribution, the powder stream, and the semi-infinite substrate together, while considering the attenuated laser power intensity distribution, the heated powder spatial distribution, and the melt pool 3D shape with its boundary variation. The intensity of the laser beam is attenuated by the powder stream during their interaction and total attenuation is a sum effect of scattering and absorption. The total heat input is, therefore, a combination of the attenuated laser intensity, $I_A$, and heated powder energy intensity, $I_P$. Here, $I_A$ is related to laser power, $P$. The temperature distribution, $T(x_0, y_0, z_0)$, where $(x_0, y_0, z_0)$ and $(\xi, \eta, \zeta)$ are moving and fixed Cartesian coordinates, respectively, is given by:

$$T(x_0, y_0, z_0) - T_0 \tag{1}$$
$$= \frac{1}{2\pi k_p} \int_{\xi=-r_L(z_0)}^{\xi=r_L(z_0)} \int_{\eta=-\sqrt{r_L^2(z_0)-\zeta^2}}^{\eta=\sqrt{r_L^2(z_0)-\zeta^2}} [\alpha_L \, I_A(\xi, \eta, z_0) + I_p(\xi, \eta, z_0)]$$
$$\times \frac{\exp\left[-\frac{v(x_0 - \eta + R)}{2\alpha_L}\right]}{R} d\eta d\xi,$$
$$R = \sqrt{(x_0 - \xi)^2 + (y_0 - \eta)^2 + z_0^2}$$

Here, $v$ is the scan velocity, $\alpha_L$ is the laser absorptivity, $k_p$ is the thermal conductivity of the powder, $r_L(z_0)$ is the effective radius of the laser beam and $T_0$ is the initial temperature of the substrate. The reader is referred to [17] for more details on the formulations of the HF model. The model is validated using experimental data for IN625 [17]. Fig. 2 compares the melt pool width predictions obtained from the model with experimental data for two cases of mass flow rates. An average deviation of 4.3% is observed between the analytical and experimental values for melt pool width which is within the acceptable limits reported in literature pertaining to AM [18], [19].

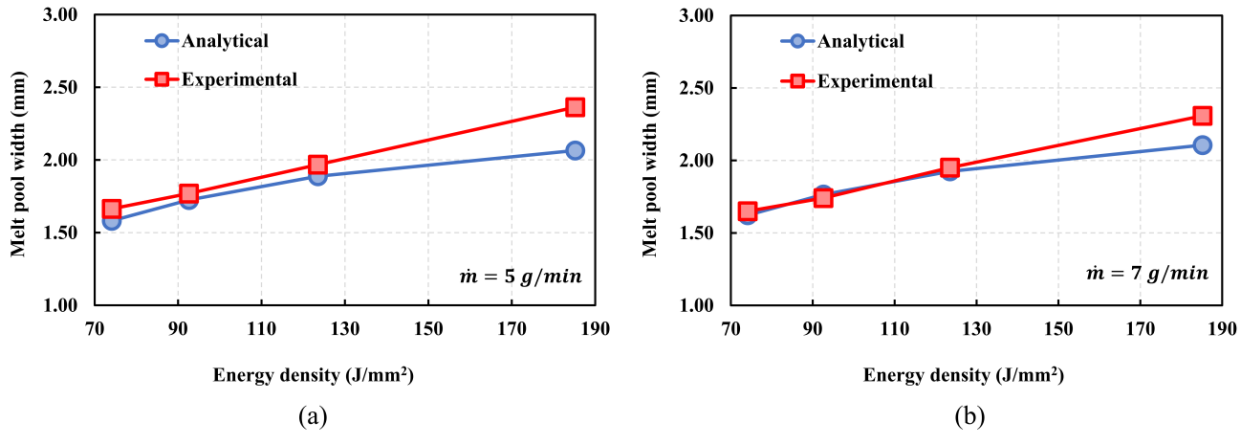

**Fig. 2. Comparison of melt pool width from experimental analysis and the HF analytical model**

prediction for IN625 with (a) $\dot{m}$ = 5 g/min, and (b) $\dot{m}$ = 7 g/min. Here, $P$ = 1000 W, $\dot{g}_{sh}$ = 2.5 dL/min, and $H$ = 7 mm. The values for other parameters can be found in reference [17].

The input space of this model is an expansive set of processing conditions and machine characteristics from which the following are identified to be easily adaptable during an L-DED process: the linear energy density controlled by laser power ($P$) and scan speed ($v$), mass flow rate of powder ($\dot{m}$), flow rate of the carrier/shielding gas ($\dot{g}_{sh}$), and nozzle height ($H$). These factors have been previously demonstrated to impact the melt pool properties individually and interactively [20], [21], [22] and, therefore, act as prospective variables in this paper. The complex nature of L-DED because of this interplay of various parameters makes the traditional DOE infeasible. Overlooking any process parameter in a modeling framework without considering its potential implications on the deposition process would, therefore, affect the prediction of as-deposited properties.

*2.1.2 Low-dimensional low-fidelity (LF) thermal model*

The thermal model developed by Eagar and Tsai is used to generate low-dimensional lower-fidelity (LF) melt pool data by querying an analytical formulation [23]. This function solves the temperature distribution produced by a Gaussian heat source over a semi-infinite domain. By ignoring convection and radiation heat transfer, the resultant temperature of the substrate is expressed as Green's function for steady-state heat conduction. For simplification, the substrate is considered quasi-steady state semi-infinite with constant thermophysical properties. The final temperature at a location $(x_0, y_0, z_0)$ and time $t$ is expressed as:

$$T(x_0, y_0, z_0) - T_0 = \frac{\alpha_L P}{\pi \rho_p c_p (4\pi a_p)^{1/2}} \int_0^t \frac{dt'(t-t')^{-1/2}}{2a_p(t-t') + \sigma_L^2} \exp\left(-\frac{(x_0 - vt')^2 + y_0^2}{4a_p(t-t') + 2\sigma_L^2} - \frac{z_0^2}{4a_p(t-t')}\right) \quad (2)$$

For the L-DED process, in the above equation, the laser power is denoted as $P$, and its corresponding scan velocity is $v$ in the $x$-direction. Other parameters pertaining to the laser include $\alpha_L$, the absorptivity coefficient, and $\sigma_L$, the standard deviation of the Gaussian profile. The properties $\rho_p, c_p$ and $a_p$ are the density, specific heat and thermal diffusivity constants for the material.

Table 1. Description of the input spaces and fidelities

| Input process parameter window | |
|---|---|
| Laser power ($P$) | 700-1000 *W* |
| Scan speed ($v$) | 5-10 *mm/s* |
| Mass flow rate of powder ($\dot{m}$) | 3-7 *g/min* |
| Nozzle height ($H$) | 3-7 *mm* |
| Carrier/Shielding gas flow rate ($\dot{g}_{sh}$) | 2-5 *dL/min* |
| **Fidelity** | **Input** | **Run time (s)** |
| Low | $x_{LF} \in \mathbb{R}^2$ | 9.484 |
| High | $x_{HF} \in \mathbb{R}^5$ | 99.963 |

The sole purpose of having the LF model is to reduce the amount of data required from the HF model to build a reliable surrogate. Hence, this LF model is calibrated to mimic the results of the HF model. While a replica cannot be generated due to the limited process parameters incorporated, this calibration attempts to tune the LF model to generate physically realistic melt pool measures for the IN625 material system, within the range of its HF counterpart. For this, the model parameters of the LF thermal model are calibrated such that the results closely match the HF model results. The variable, $\sigma_L$ of Eq. 2 is considered as $\phi \times r_L$. Here, $\phi$ is left as a calibration parameter while $r_L$ is machine-specific and shared across both thermal models. Furthermore, both models share mainly three parameters - $P$, $v$, and $r_L$ of which, only $P$ and $v$ are user-modifiable process parameters. Calibration of the LF model uses a full factorial DOE evaluated using the HF model with three levels each for $P$ (700, 850, and 1000 W) and $v$ (5, 7.5, and 10 mm/s). The three inputs absent in the LF model: $\dot{m}$, $H$, and $\dot{g}_{sh}$, are kept at the mean value of their respective ranges (Table 1). The constant material properties of IN625 and the inputs for the HF thermal model, excluding those considered in the DOE, are selected based on the experimental validation presented in the literature [17]. This data is used to calibrate the LF model by tuning $\phi$ and $\alpha_L$ to obtain steady-state melt pool depth ($\delta$) and width ($\omega$) estimates close to the HF equivalents. Both models are developed on MATLAB® and their solution time on an Intel® Xeon® 256 GB RAM is reported in Table 1. The HF model is slower than the LF model by 11 times.

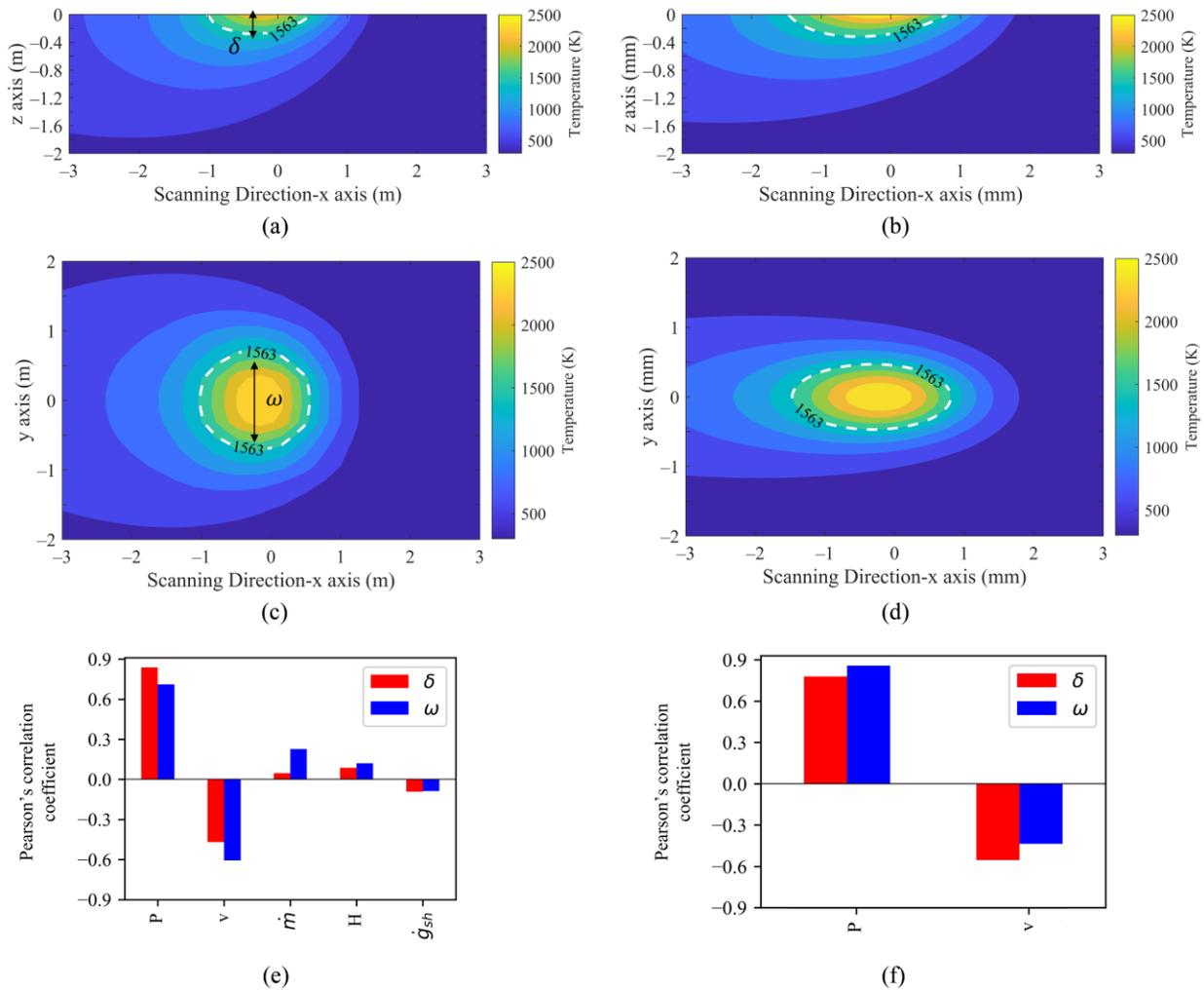

**Fig. 3.** Calibration results for the lower-fidelity thermal model. The temperature contours calculated using the HF model with the (a) melt pool depth (δ) marked in the x-z cross-section, and (c) melt pool width (ω) marked in the x-y cross-section. (e) Pearson's correlation coefficients calculated between the inputs to the HF model and its outputs, δ and ω. Similarly, the temperature contours calculated using the calibrated LF model with the (b) melt pool depth (δ) marked in the x-z cross-section, and

**(d) melt pool width ($\omega$) marked in the x-y cross-section. (f) Pearson's correlation coefficients calculated between the inputs to the LF model and its outputs, $\delta$ and $\omega$.**

Fig. 3(a-d) shows a representative case for melt pool cross-sections obtained with $P$ = 850 W and $v$ = 7.5 mm/s. The LF model results in a smaller elongated melt pool compared to the HF model. The thermal gradients in the LF model are also much higher than their counterpart. These variations arise due to the difference in how the heat input is considered. The LF model only considers the attenuated laser intensity whereas the HF model couples the attenuated laser beam and heated powder stream at an incline. The $R^2$ values after calibration for $\delta$ and $\omega$ are 0.99 and 0.96, respectively, within the DOE used. Pearson's correlation coefficients are calculated for the HF and LF responses in Fig. 3(e) and (f). Since the fundamental energy conservation laws remain true for both models, the correlations of the melt pool dimension with the common input parameters ($P$ and $v$) remain similar. Within the process parameter window considered, $P$ has a greater impact on $\delta$ and $\omega$ than $v$. Weaker positive correlations are observed for $H$ and $\dot{m}$ and an inverse correlation in melt pool dimensions is observed with $\dot{g}_{sh}$.

## 2.2 Data generation for Het-MFGP

Table 1 presents the five main process parameters selected for the HF model along with their ranges for the DOE considered in this paper. The LF inputs ($P$ and $v$) are a subset of the HF input space, varying within the same ranges as prescribed in Table 1. Throughout the paper, LHS is employed to generate training data within the input spaces of the respective models. Both HF and LF models are queried at these input points which together with the output form the training data for MF modeling. The test data is kept fixed. This comprises another random LHS performed within the HF input space to obtain 100 data points.

## 2.3 Development of the Het-MFGP framework

Multi-fidelity surrogates combine data from different models by statistically learning correlations within the data, this approach yields a precise and efficient system approximation [24]. Co-kriging approaches have been employed to merge the outputs from these correlated variables [24], [25], [26]. In this study, an autoregressive (AR1) version of Kennedy and O'Hagan's co-kriging method [3] is utilized. The co-kriging approach involves constructing individual surrogate models for different fidelity levels and linking them together using a covariance function within a GP framework. For a given set of input process parameters denoted as $x$, the auto-regressive representation of the LF and the HF models at two fidelity levels can be expressed as follows:

$$y_{HF} = \rho y_{LF} + \gamma_{HF}(x) \tag{3}$$

Each fidelity model in AR1 is assigned a GP prior. Here, $y_{LF}$ is the LF prior and $y_{HF}$ is the HF prior. The parameter, $\rho$ quantifies the correlation between $y_{HF}$ and $y_{LF}$ and is assumed as a constant function in this paper. The $\gamma_{HF}(x)$ in Eq. 3 represents the discrepancy between $y_{LF}$ and $y_{HF}$. Here, $\gamma_{HF}(x)$ and $y_{LF}$ are two independent Gaussian processes given by:

$$\gamma_{HF}(x) \sim GP(0|k_{HF}(x,x')) \tag{4}$$

and

$$y_{LF} \sim GP(0|k_{LF}(x,x')) \tag{5}$$

In the above equations, $x'$, corresponds to an input other than $x$. This means,

$$y_{HF} \sim GP(0\,|\rho^2 k_{LF}(x,x') + k_{HF}(x,x')) \tag{6}$$

The joint Gaussian distribution takes the form:

$$\begin{bmatrix} y_{LF} \\ y_{HF} \end{bmatrix} \sim GP\left(\begin{bmatrix} 0 \\ 0 \end{bmatrix}, \begin{bmatrix} k_{LF}(x,x') & \rho k_{LF}(x,x') \\ \rho k_{LF}(x,x') & \rho^2 k_{LF}(x,x') + k_{HF}(x,x') \end{bmatrix}\right) \quad (7)$$

Here, $k_{HF}$ and $k_{LF}$ are the kernel functions corresponding to the HF and LF models, respectively. For simplicity, in this paper, the kernels $k_{HF} = k_{LF} = k$, and are specifically the Matérn 5/2 kernel.

In this study, the HF and LF models differ in dimensionality, preventing direct combination using Eq. 7. Specifically, the HF thermal model involves five input parameters, while the LF model only has two. To address this, the HF model's input space is aligned with that of the LF model through a parametrized mapping function. This mapping of the HF input space onto the LF input space is calibrated using IMC [4] such that it minimizes the difference between the LF and HF outputs. Unlike other mapping techniques used for heterogeneous domains in literature, such a calibration can be applied to cases where the LF input space is a subset of the HF input space. Let $x_{HF}$ and $x_{LF}$ correspond to a vector in $X_{HF}$ and $X_{LF}$, which are the input spaces of the HF and LF model, respectively, such that $X_{HF} \in \mathbb{R}^{d_{HF}}$ and $X_{LF} \in \mathbb{R}^{d_{LF}}$. Here, $d_{HF}$ is the dimension of the HF input space and $d_{LF}$ is that of the LF input space where $d_{HF} > d_{LF}$. A linear transformation $g(\cdot)$ is defined to map the HF input to the LF input such that:

$$y_{HF}(x_{HF}) = y_{LF}(g(x_{HF}, \beta)) + \mathcal{L} \quad (8)$$

$$g(x_{HF}, (A, b)) = A x_{HF} + b \quad (9)$$

Here, $A$ is a $d_{LF} \times d_{HF}$ matrix and $b$ is a $d_{LF} \times 1$ vector composed of elements from the calibration parameter vector, $\beta$. To calculate the calibration parameters, the squared loss $\mathcal{L}$ must be minimized.

$$\mathcal{L}(\beta) = \sum_{i=1}^{N_{HF}} \left( y_{HF,i} - y_{LF}\left(g(x_{HF,i}, \beta)\right) \right)^2 \quad (10)$$

A regularization parameter, $\lambda$ is introduced to minimize overfitting. Regularization is introduced with nominal values for $\beta$ denoted by $\beta_0$. The nominal mapping is the identity mapping of $\mathbb{R}^{d_{HF}}$ from $\mathbb{R}^{d_{LF}}$. The elements of $\beta_0$ yield the nominal parameters of $A$ and $b$ given by $A_0$ and $b_0$. Using 2-norm regularization, the loss function becomes:

$$\underbrace{\mathcal{L}(A,b)}_{\text{Loss function}} = \underbrace{\sum_{i=1}^{N_{HF}} \left( y_{HF,i} - y_{LF}(A x_{HF,i} + b) \right)^2}_{\text{Squared-loss}} + \underbrace{\lambda}_{\text{Regularization parameter}} \underbrace{(\|A - A_0\|_2 + \|b - b_0\|_2)}_{\text{Regularization term}} \quad (11)$$

Fig. 4 details the heterogeneous domain mapping. For HF, the dimension of the input $X_{HF}$, is $d_{HF} \times N_{HF}$, where $N_{HF}$ is the number of HF input points. The corresponding output is given by $y_{HF}$. Similarly, the quantity of the LF dataset is defined by $N_{LF}$ ($>N_{HF}$). The map $g(X_{HF}, \beta)$ transforms $X_{HF}$ into $G$ of dimension $d_{LF} \times N_{HF}$ such that the predictions by the LF model for the inputs $G$ is close to the actual $y_{HF}$. This is realized by the loss function $\mathcal{L}$ which is minimized by optimizing $\beta$. These details are further outlined in Algorithm 1

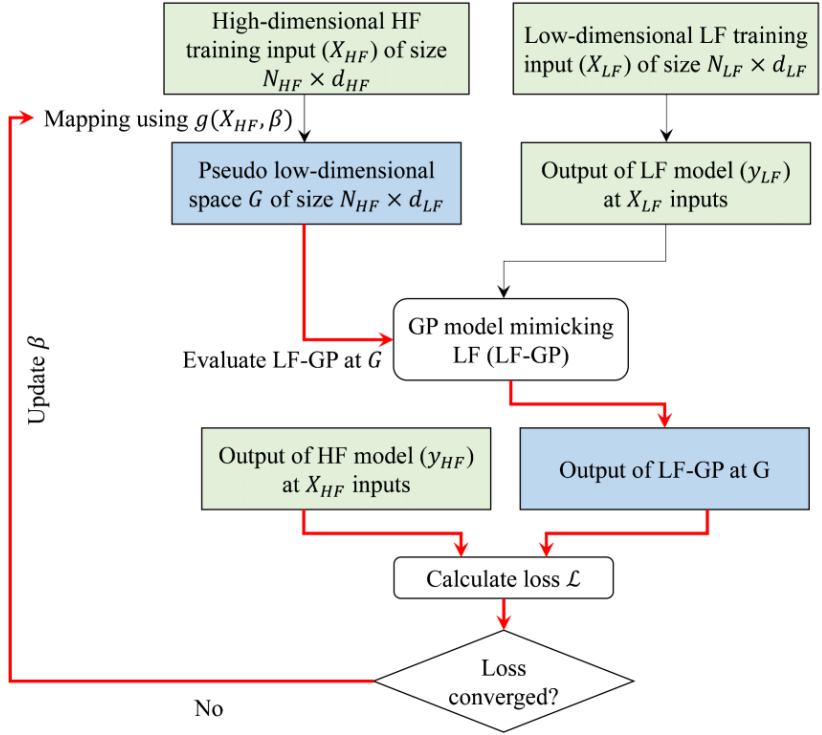

**Fig. 4.** The heterogeneous domain mapping to prepare the high-dimensional, HF data for co-kriging using IMC. The original inputs and outputs are colored green and new inputs and outputs generated during the IMC process are colored blue. The red arrows guide the calibration process until convergence. The HF and LF datasets are $X_{HF}$ mapping to $y_{HF}$ and $X_{LF}$ mapping to $y_{LF}$, respectively. The size of the $X_{HF}$ is denoted as $N_{HF} \times d_{HF}$, where $N_{HF}$ is the number of HF training data points and $d_{HF}$ is the dimension. Similarly, the size of the $X_{LF}$ is denoted as $N_{LF} \times d_{LF}$, where $N_{LF}$ is the number of LF training data points and $d_{LF}$ is the dimension. The mapping, $g(X_{HF}, \beta)$, transforms $X_{HF}$ to $G$ in $\mathbb{R}^{d_{LF}}$ such that this pseudo low dimensional space is of size $N_{HF} \times d_{LF}$. A GP surrogate, LF-GP, is trained on the LF data $(X_{LF}, y_{LF})$. This GP is used to calculate outputs at $G$. The $\beta$ parameter is optimized to minimize the loss, $\mathcal{L}$, which is a measure of the distance between $y_{HF}$ and the LF-GP's output for $G$.

---

**Algorithm 1** Algorithm for mapping the heterogeneous domain for Het-MFGP

1:    **Data**: $f_{HF}: X_{HF} \rightarrow y_{HF}$, $f_{LF}: X_{LF} \rightarrow y_{LF}$
2:    Initialize $\beta$
3:    Initialize $i = 0$
4:    **while** $i < n_{iter} < n_{iter}$ **do**:                                                     ▷ $n_{iter}$: Max. iterations for optimization
5:       $\beta = [A, b]$
6:       $G = AX_{HF} + b$
7:       Calculate $\mathcal{L}(\beta)$
8:       **if** $\mathcal{L}(\beta) > \varepsilon$ **then**:                                                        ▷ $\varepsilon$: convergence tolerance
9:          Update $\beta$
10:         $i++$
11:      **else**
12:         break
13:      **end if**
14:   **end while**

## 2.5 Performance Metrics

The selection and performance of the surrogate models are assessed by employing the following metrics common to regression tasks:

- $R^2$: The coefficient of determination calculated as $1 - \frac{\sum_i^N (y_i - \hat{y}_i)^2}{\sum_i^N (y_i - \bar{y})^2}$ where $y$ is the vector of true values for $N$ data points whose mean is $\bar{y}$ and $\hat{y}$ corresponds to the mean of the predictive posterior. A high $R^2$ indicates that the model is a good fit for the data and can explain most of the observed variation.

- $L2$: The relative L2 error norm calculated as $\frac{\|y - \hat{y}\|_2}{\|y\|_2}$, where $\|\cdot\|_2$ is the L2 norm. A lower $L2$ signifies that the model's predictions are closer to the true values, i.e., the model's predictions have less average deviation from the actual data points.

- $\sigma_{avg}$: The average standard deviation which is calculated $\sqrt{\sum_i^N \sigma_i^2 / N}$ where $\sigma_i^2$ is the variance associated with the data point indexed $i$ in a data set of $N$ data points. The value of $\sigma_{avg}$ is a measure of model uncertainty where a lower value is indicative of higher confidence in the predictions.

If any of the above metrics are associated with an overline (·), it signifies that these metrics are averaged across multiple iterations. In this paper, specifically ten iterations are used to avoid any bias induced by the initial DOE used for training the surrogates.

## 3 Results and discussion

The characterization of melt pool geometry is performed by separate predictive Het-MFGP surrogates for $\delta$ and $\omega$, each. These surrogates are calibrated for heterogeneous domain mapping and surrogate accuracy. The performance of the Het-MFGP surrogates is then evaluated using relevant metrics against a baseline single-fidelity GP. The HF model's outputs are considered as the ground truth for these test data points. Finally, sensitivity analyses are conducted to determine which process parameters, or combinations of them, have the most significant impact on the surrogate.

### 3.1 Optimizing Het-MFGP for accuracy

Optimization for the heterogeneous map, i.e., $g$ is carried out by calibrating the parameter $\beta$ in Eq.10 such that the loss, $\mathcal{L}$ is minimized. The nominal parameters of $\beta_0$ - $A_0$ and $b_0$ are an identity matrix and zero-vector of the same dimensions as $A$ and $b$, respectively. In total, there are $(d_{LF} \times d_{HF}) + d_{LF} \Rightarrow (2 \times 5 + 2 = 12)$ parameters for calibration. To limit the computational expense of the optimization, the calculation of the objective function, as defined in Eq.11, uses a GP surrogate trained on the LF data available for training. From a practical standpoint, the use of such a surrogate avoids the need to reevaluate the LF model repeatedly during the calculation of $\mathcal{L}$. For only 20 LF data points, this GP surrogate scores a 5-fold cross-validation score of $R^2 = 0.99$ for both $\delta$ and $\omega$. Fig. 5 shows convergence results for an illustrative case of the optimization process for $N_{HF} = N_{LF} = 20$. The optimizer's exploration of the space can be visualized by observing the distance between consecutive evaluations ($n$ and $n - 1$), as shown in Fig. 5(a). This distance, $\| \beta_n - \beta_{n-1} \|_2$, is plotted over the $n$ iterations incurred during optimization. Often, there are noticeable intervals between evaluations; however, occasionally, consecutive evaluations seem to occur in close proximity. These closely sequenced evaluations typically align with a decrease in the value of the best-selected sample. Fig. 5(b) shows the convergence of the optimizer to the best value of $\beta$ for which $\mathcal{L}$ is minimized. The objective function or the $\mathcal{L}$ steadily decreases over iterations, indicating convergence towards the optimal solution is achieved within ~50 iterations (65.87 secs).

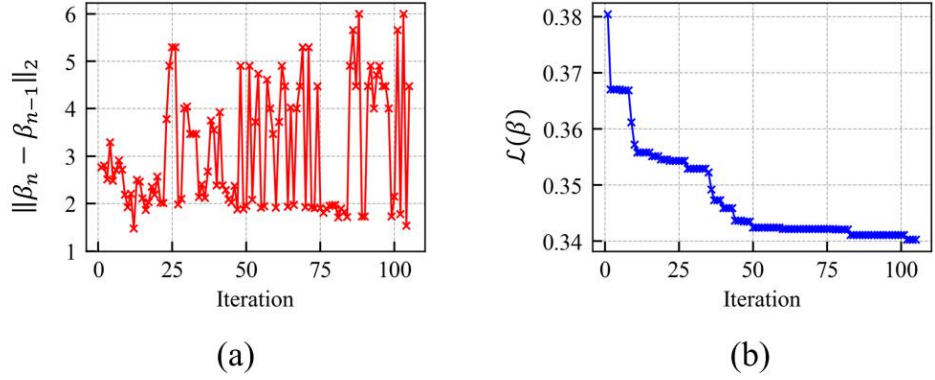

**Fig. 5. Optimization results for calibration of $\beta$ for heterogeneous domain mapping. (a) The distance between the last two observations ($n$ and $n-1$) denoted as $\|\beta_n - \beta_{n-1}\|_2$ over the iterations incurred in the optimization process. (b) The convergence plot with the loss $\mathcal{L}$ at the best location previous to each iteration plotted against the iteration.**

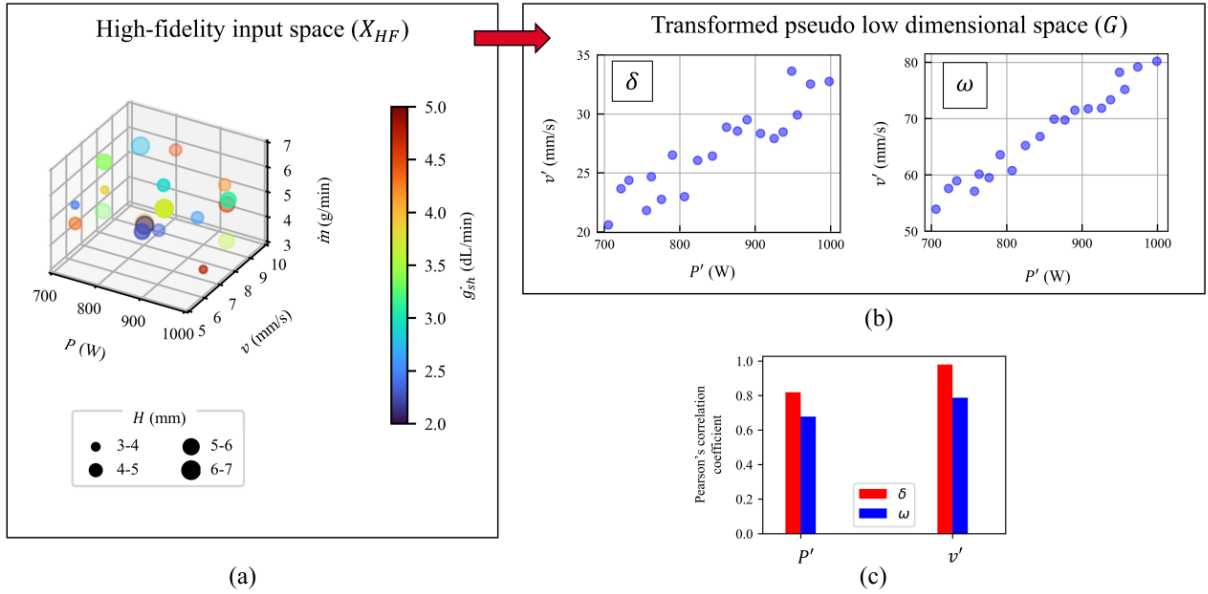

**Fig. 6. Visualization of the HF input space (a) before transformation via heterogeneous domain mapping in the $\mathbb{R}^5$ space, and (b) after applying the heterogeneous domain map to the $\mathbb{R}^2$ space. (c) Pearson's correlation coefficients calculated for $\delta$ and $\omega$ with the new pseudo inputs $P'$ and $v'$.**

Fig. 6(a) shows the original input space of the HF model, for an LHS DOE of 20 points on a 3D-scatter plot with $P$, $v$, and $\dot{m}$ along the three axes. The points are colored according to the value of $g^*_{sh}$ and sized based on their $H$ value. The mapping $g: \mathbb{R}^5 \rightarrow \mathbb{R}^2$ is learned to ultimately lead to the space, $G$ as shown in Fig. 6(b), for $\delta$ and $\omega$. The mapping yields a combination of new inputs suppressed into the low-dimensional space that can be assumed as the pseudo laser power, $P'$ and scan velocity, $v'$. The scatter plot in Fig. 6(b) reveals that the values for $P'$ remain relatively unchanged compared to $P$, while the effects of the remaining input parameters merge into the new $v'$. Notably, $v'$ exhibits values higher than the original ranges for $v$. Comparing the mappings observed for $\delta$ and $\omega$, it is evident that the latter results in greater values for $v'$. This can be attributed to the comparatively stronger influence of other process parameters on $\omega$ as seen in Fig. 3(e) whose combined effect is encapsulated within $v'$. This observation is further corroborated by the Pearson's correlation coefficients recalculated for the new inputs $P'$ and $v'$ where positive correlations are observed for both the outputs, as depicted in Fig. 6(c).

## 3.2 Development of the Het-MFGP surrogate

The accuracy of the Het-MFGP surrogate can be affected by three parameters: (i) the number of HF data points for training, $N_{HF}$, (ii) the number of LF data points for training, $N_{LF}$, and (iii) the 2-norm regularizer, $\lambda$. The values for $N_{HF}$ and $N_{LF}$ depend on the resources available and the accuracy/confidence necessary. Essentially, a well-calibrated Het-MFGP surrogate aims to strike the right balance between the two types of data. The HF data points, which are more computationally expensive but yield precise results, are judiciously used alongside LF data points, which are less accurate but computationally cheaper. The challenge lies in determining the optimal number of HF data points needed to refine the surrogate's accuracy and the appropriate number of LF data points to ensure efficiency without compromising the overall performance. Striking this balance is crucial to harness the full potential of Het-MFGP. Data size that is ten times the input dimension is considered canonical while modeling HF data [27]. This necessitates a minimum of $N_{HF} = 50$ for building a GP surrogate that can emulate the HF model and give reliable estimates of melt pool dimensions. For this study, the number of test data points is kept constant at 100 which are obtained by a random LHS within the HF process parameter window. Since the effects of varying $N_{HF}$ and $N_{LF}$ are studied for this fixed quantity of test data points, the observations can be used to make decisions on relative quantities of data when the data size for testing increases.

Fig. 7 shows the $\overline{L2}$ errors and $\overline{\sigma_{avg}}$ values for surrogates developed to predict $\delta$ and $\omega$. The study examines scenarios where $N_{HF}$ is set to 5, 10, 20, and 30, while $N_{LF}$ varies from 0 to 40 in increments of 10. The deliberate choice of keeping $N_{HF}$ below 50 serves the purpose of evaluating the performance of the surrogate in scenarios with limited available data. When $N_{LF}$ is set to 0, the GP surrogate is trained only on HF data points. As expected, the errors and uncertainty reduce for both $\delta$ and $\omega$ with increase in $N_{HF}$. The positive impact of incorporating LF points is evident in terms of improving the $\overline{L2}$ values until $N_{LF} = 20$. Beyond this threshold, a marginal increase in $\overline{L2}$ error is observed, attributed to the introduction of bias by the LF data points to the surrogate. The $\overline{\sigma_{avg}}$ continues to reduce with an increase in $N_{LF}$ for surrogates trained on both $\delta$ and $\omega$ data. An anomaly is observed with the lowest $\overline{\sigma_{avg}}$ for a surrogate trained on just $N_{HF} = 5$. This peculiar case may be attributed to chance initializations of HF training data closer to the actual test points, despite reporting average values over multiple iterations. Based on the comprehensive analysis of both metrics, $N_{HF} = N_{LF} = 20$ is selected as the optimal configuration for training a reliable Het-MFGP surrogate capable of accurately predicting melt pool dimensions. The relatively modest improvements in accuracy despite increasing $N_{HF}$ suggest that beyond a certain point, the returns from additional HF data diminish. Furthermore, the results indicate that the contribution of LF data becomes particularly valuable when $N_{LF}$ is kept at an intermediate value, such as 20. This observation indicates that, with an optimal blend of HF and LF data, the surrogate model can effectively leverage the complementary strengths of both types of data.

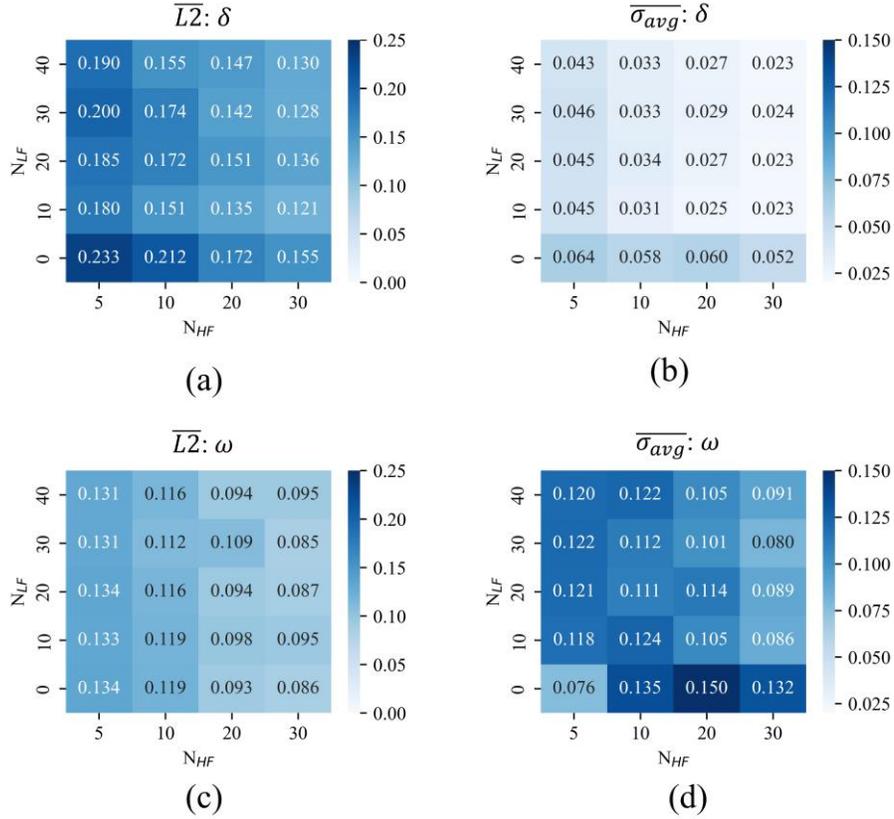

**Fig. 7. Evaluating the accuracy of Het-MFGP surrogate by varying the number of HF data points ($N_{HF}$) and the number of LF data points ($N_{LF}$). The metrics calculated on the test data of 100 HF points are averaged over ten initializations of the $N_{HF}$ and $N_{LF}$ for training the surrogates. The $\overline{L2}$ values are shown in (a) and (c) for $\delta$ and $\omega$ and the $\overline{\sigma_{avg}}$ are shown in (b) and (d) for $\delta$ and $\omega$.**

Once $N_{HF}$ and $N_{LF}$ are fixed, the regularization parameter, $\lambda$ is calibrated to optimize the heterogeneous mapping to improve the accuracy of the surrogate. Four cases of $\lambda$ of varying orders of magnitude, [0.001, 0.01, 0.1, 1] are considered. A lower value of $\lambda$ promotes faster optimization albeit with a tendency to overfit. For the current optimization challenge, an increase in $\lambda$ by an order of magnitude roughly corresponds to 20 additional iterations for convergence. The data presented in Fig. 8 illustrates the averaged metrics, $\overline{L2}$ and $\overline{\sigma_{avg}}$, calculated for surrogates calibrated with multiple values of $\lambda$ for $\delta$ and $\omega$. The results reveal that the Het-MFGP surrogate trained for $\delta$ records a higher $\overline{L2}$ but a lower $\overline{\sigma_{avg}}$ compared to that trained on $\omega$. This bias-variance interplay implies that, on the whole spectrum of $\lambda$ values, the inclination towards overfitting is more pronounced for $\omega$ data points than for their $\delta$ counterparts. This observation aligns with the patterns evident in Fig. 7. No substantial variation in performance is observed with change in $\lambda$ for the surrogates trained for $\omega$. Regarding the selection of $\lambda$, opting for $\lambda = 0.01$ strikes a suitable balance between metrics while demanding fewer optimization steps than with higher values of $\lambda$.

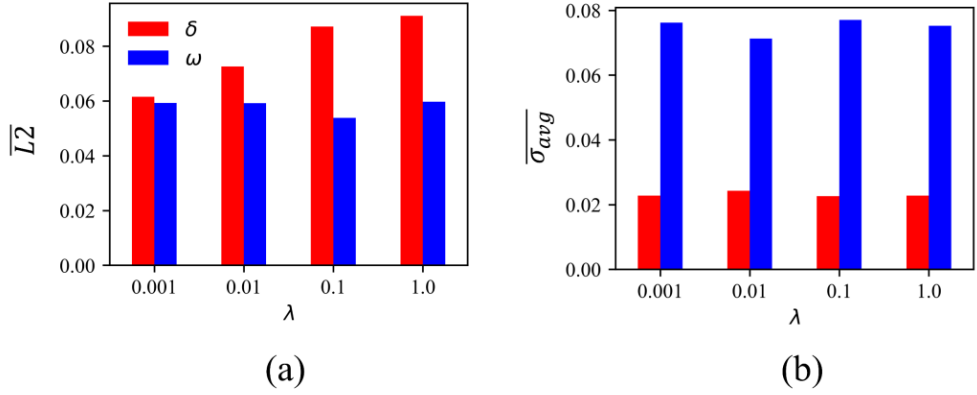

**Fig. 8. Results from the calibration of the surrogate for accuracy by varying $\lambda$. The averaged metrics for different orders of magnitude of $\lambda$ for the two surrogates are shown in (a) for $\overline{L2}$ and (b) for $\overline{\sigma_{avg}}$.**

### 3.3 Predictive capability of the Het-MFGP surrogate

The calibrated Het-MFGP surrogate's posterior distribution is used to make predictions on the unseen data points. The mean of the posterior distribution gives the point estimate of the prediction, while the covariance gives a measure of uncertainty. Separate Het-MFGP surrogates are trained and tested for $\delta$ and $\omega$, which are discussed in Fig. 9 and 10, respectively. For establishing a baseline value for the performance metrics, the Het-MFGP surrogate is compared with a simple GP's prediction where this single-fidelity surrogate is trained on an equivalent amount of HF data.

Fig. 9(a) and (b) show the parity plots with the observed values for melt pool depth ($\hat{\delta}$) from the single-fidelity and Het-MFGP surrogates plotted against the true value ($\delta$), respectively. Fig. 9(a) displays the GP's prediction for $\delta$, which is trained on 20 HF data points and tested on the LHS DOE of 100 data points. The GP's predictions are scattered around the HF ground truth and record an $R^2$ of 0.592 with a $\sigma_{avg}$ of 0.059. Based on the observations from Fig. 7 in Section 3.2, adding LF data points is expected to improve the quality of predictions. Fig. 9(b) shows the Het-MFGP counterpart trained and tested on the same HF data points as the GP in (a) while being augmented with 20 LF data points and tested on the same LHS DOE of 100 HF data points. The Het-MFGP is successful in making predictions of higher accuracy in the test domain with an $R^2$ of 0.975 and $L2$ error of 0.041 which is lower than its GP counterpart by an order of magnitude. Despite the discrepancy between the LF model and the HF ground truth, it improves the surrogate's accuracy by significantly reducing its prediction scatter. Even in the presence of limited HF ground truth and heterogeneous input domains between the fidelities, the Het-MFGP surrogate reduces the $\sigma_{avg}$ of prediction by 3 times, from 0.059 to 0.019.

Fig. 9(c) visualizes the errors observed in both surrogates as a histogram plot. The errors for both GP and Het-MFGP surrogates follow a normal distribution with the latter having a smaller standard deviation. It is inferred that 95% of the errors (calculated as the difference between the observed and true values) lie within the bins that correspond to $-0.1$ to $0.1$. Fig. 9(d) reports Sobol sensitivity indices to quantify the contribution of each input variable to the overall variance in Het-MFGP's predictions. Here, values for $S1$ and $ST$ indices which constitute the first-order (individual contribution) and total (individual and higher-order sensitivities arising from interactions) sensitivities, respectively are provided. These indices bring forth a quantitative measure of sensitivity and can guide the prioritization of inputs for further analysis or refinement. The inputs $P$ and $v$ clearly dominate the rest by accounting for $\sim 0.9$ in just $S1$ sensitivity indicating that very little uncertainty is introduced by $\dot{m}$, $H$, and $\dot{g}_{sh}$ or any pairwise or higher order interactions between the inputs.

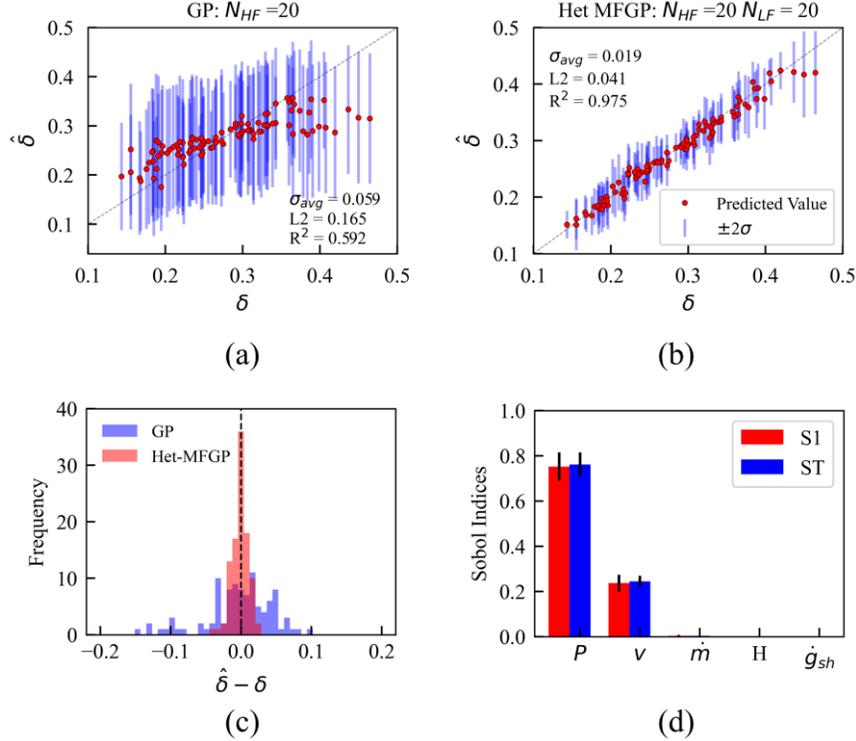

**Fig. 9. Parity plot comparing ground truth, $\delta$, and predicted values, $\hat{\delta}$, for the test inputs as predicted by (a) the single fidelity GP surrogate model learned only from 20 HF data, and (b) the Het-MFGP surrogate learned on 20 HF data and 20 LF data. The $\hat{\delta}$ is plotted on the y-axis against the $\delta$ on the $x$-axis. The predicted values are indicated by the red circles along with the $\pm 2\sigma$ bars in blue. (c) Histograms of error in $\hat{\delta}$ w.r.t $\delta$ as predicted by surrogate models learned only from the Het-MFGP surrogate and single fidelity GP surrogate model learned only from HF data (d) Sobol sensitivity indices - first order: $S1$ and total order: $ST$, of the inputs.**

Fig. 10 shows the results obtained for predictions of $\omega$ tested on the same input points as those employed for validating the surrogates for prediction of $\delta$. The Het-MFGP surrogate, in this case as well, improves the $L2$ error, reducing it from 0.09 to 0.034 with the addition of 20 LF data points. Both models do not exhibit bias as evinced by the predictions that are randomly scattered about the $y = x$ line on the parity plot. The GP surrogate, however, over predicts the smaller values of $\omega$ and under predicts for larger $\omega$ values. The scatter is minimized along with shorter $\pm 2\sigma$ bars in Fig. 10(b) for Het-MFGP where the $\sigma_{avg}$ is reduced from 0.154 to 0.063. The histogram of errors for the GP and Het-MFGP surrogates is plotted in Fig. 10(c). The errors calculated for the Het-MFGP and GP surrogate follow a normal distribution with the latter's errors being spread out over a larger range. 95% of the errors for the Het-MFGP surrogate lie within the bins corresponding to $-0.2$ and $0.2$. On the contrary, the errors incurred by the GP surrogate are distributed across the range of bin values depicted in the plot. Compared to $\delta$, the uncertainty observed for the $\omega$ predictions are marginally higher for both the GP and Het-MFGP surrogates. Although the LF model can produce values for $\delta$ that are close to the actual values, the discrepancy with the HF model during the calibration itself was higher for $\omega$. Additionally, it is important to recognize that this calibration solely involved HF estimates with variation in $P$ and $v$, neglecting the potential impact of the other three powder-related parameters. Variation of these parameters, as observed in the test input points, would thus further exacerbate the deviation in predicting $\omega$. The Het-MFGP surrogate's attempts to compensate for the differences between the two fidelities, ultimately, affect the accuracy and uncertainty. The sensitivities for each input that are calculated as Sobol indices are plotted in Fig. 10(d). Interestingly, the $S1$ and $ST$ indices indicate that similar to Fig. 9(d) for $\delta$, the surrogate for $\omega$ also appears to only be affected by $P$ and $v$. Despite the marginal correlations exhibited in Fig. 3(e), the contribution by inputs $\dot{m}$, $H$ and $\dot{g}_{sh}$ towards the Het-MFGP surrogate's uncertainty is almost negligible.

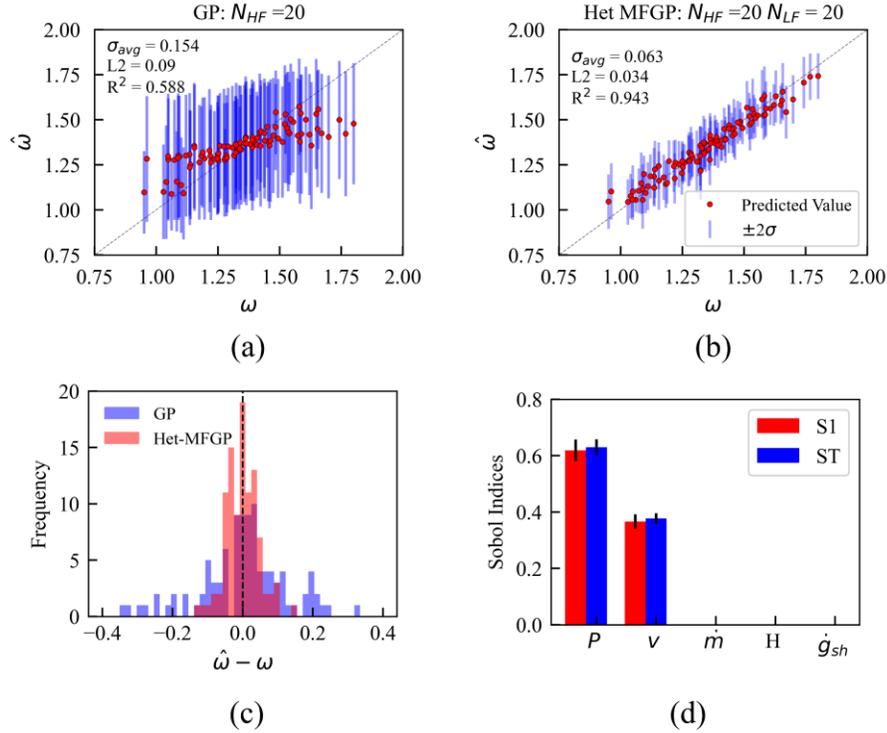

**Fig. 10.** Parity plot comparing ground truth, $\omega$, and predicted values, $\hat{\omega}$, for the test inputs as predicted by (a) the single fidelity GP surrogate model learned only from 20 HF data, and (b) the Het-MFGP surrogate learned on 20 HF data and 20 LF data. The $\hat{\omega}$ is plotted on the y-axis against the $\omega$ on the x-axis. The predicted values are indicated by the red circles along with the $\pm 2\sigma$ bars in blue (c) Histograms of error in $\hat{\omega}$ w.r.t $\omega$ as predicted by surrogate models learned only from the Het-MFGP surrogate and single fidelity GP surrogate model learned only from HF data (d) Sobol sensitivity indices - first order: $S1$ and total order: $ST$, of the inputs.

Combining results from both the Het-MFGP surrogates, a resultant melt pool can be visualized as a lower-fidelity approximation in the form of a semi-ellipse. The predicted $\hat{\delta}$ and $\hat{\omega}$ are used to define the axes of the ellipse as shown in Fig. 11. Eight combinations of inputs from the test data are randomly selected to show representative results. The shaded yellow region corresponds to the approximated melt pool using true values for $\delta$ and $\omega$. For every case demonstrated here, the true melt pool lies within the $\pm 2\sigma$ curves that are close to the predicted melt pool curve (dashed blue curve) indicating high confidence with minimal deviation.

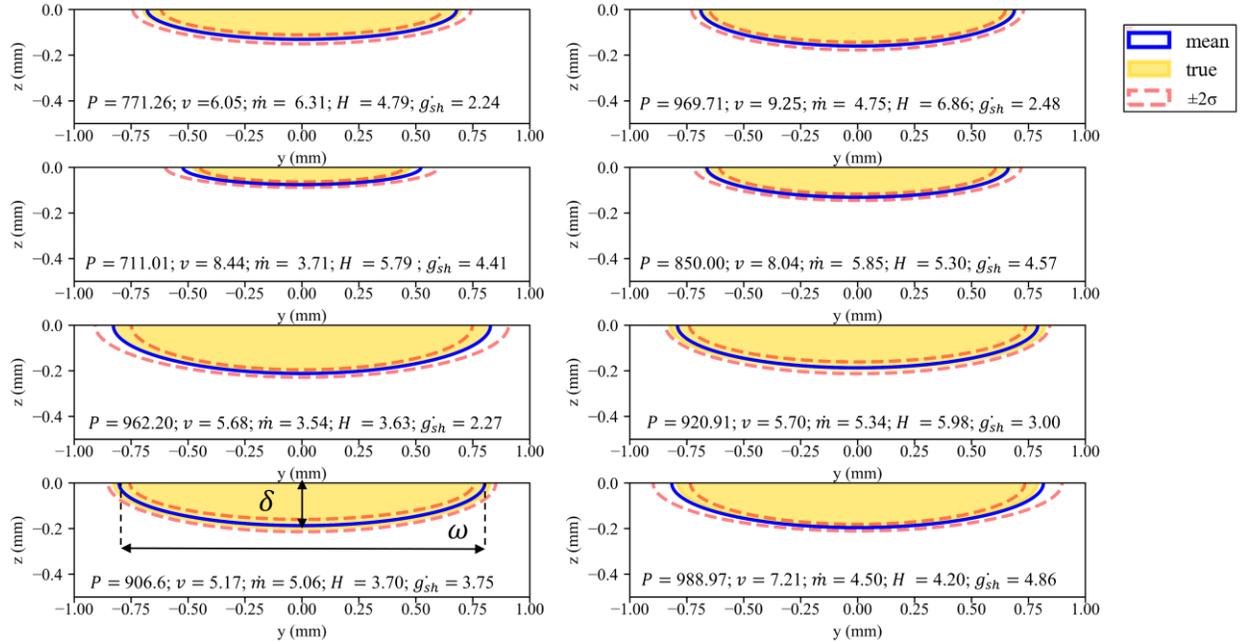

**Fig. 11. The predicted melt pool boundaries superposed on the actual HF cross-sections for eight representative cases. The actual melt pool is highlighted in yellow. The inputs associated with each case are provided in each subfigure. Here, the predicted boundary (in blue) is approximated as a semi-ellipse whose minor axis is the predicted melt pool depth and the major axis is 1/2 of the predicted melt pool width. The dashed red line corresponds to $\mu + 2\sigma$ and $\mu - 2\sigma$, where $\sigma$ is the standard deviation of the posterior.**

### 4. Conclusions

In the context of L-DED, accurately predicting melt pool geometry is crucial for effective process development as it directly influences the final properties. Melt pool models serve as important tools to obtain faster estimates of the melt pool geometry. However, existing models pose a challenge, as they vary in complexity. This enforces a trade-off between accuracy and cost when a single melt pool model is used in isolation. Additionally, many reduced-order alternatives are limited to specific dimensional spaces, necessitating compromises on the range of process parameters that can be explored. To address these challenges, this paper introduces Het-MFGP, a novel multi-fidelity Gaussian process that employs heterogeneous domain mapping. The MF aspect handles the information from different fidelities via co-kriging using GPs which ensures uncertainty quantification with minimum data requirement. The heterogeneous domain mapping takes care of the different input spaces before co-kriging.

Here, two distinct analytical melt pool models are considered, each operating on separate input spaces: the LF model focuses on laser power and scan velocity, while the HF model considers powder flow rate, nozzle height, and shielding gas flow rate. The mapping transforms the high-dimensional space into a pseudo space with the same dimension as the LF by minimizing the distance between the two models. A thorough calibration is conducted to determine the relative quantities of HF and LF data. The resulting Het-MFGP surrogate, developed using transformed HF data along with LF data, significantly enhances predictions. For 100 unseen data points in the HF domain, the Het-MFGP surrogate, augmented with 20 LF points, reduces error by an order of magnitude and uncertainty by three-fold compared to a single-fidelity GP. Notably, the tested Het-MFGP surrogates achieve high $R^2$ scores of 0.975 and 0.943 for predicting melt pool depth and width, respectively. While this study focuses on thermal models for L-DED, the proposed framework has broader applicability to predict process-structure-property relationships in metal AM processes.

The findings highlight the advantages of using the Het-MFGP surrogate for modeling melt pool behavior in AM processes. Nevertheless, further research may be needed to validate the approach in other domains or with different types of models and multimodal data. The wider effectiveness of this approach when integrated with optimization routines such as Bayesian optimization also needs to be ascertained. Additionally, here, the IMC considers a linear recursive map which while simple can be restrictive. A better alternative might be to use non-linear transformations which introduce more complexity and flexibility in capturing intricate relationships within the data. While non-linear functions can be employed provided some prior understanding of the data, a more agnostic approach would be to employ multilayer perceptron or deep neural networks which can introduce non-linear transformations through multiple layers and activation functions [28].

**Acknowledgments**
The authors would like to thank Dr. Sudeepta Mondal (Penn State) for his help during the conceptualization phase of this work. The authors would also like to thank Dr. Yuze Huang (Lancaster University) for their help with the higher-fidelity analytical model.

**Funding**
The work reported in this paper is funded in part by the Pennsylvania State University, PA 16802, USA, and, in part by the U.S. Army Engineer Research and Development Center through Contract Number W912HZ21C0001. Any opinions, findings, and conclusions in this paper are those of the authors and do not necessarily reflect the views of the supporting institutions.

**Authors' Contribution**
Nandana Menon: Methodology, Software, Validation, Formal analysis, Investigation, Data curation, Writing – original draft, Writing – review & editing, Visualization. Amrita Basak: Conceptualization, Resources, Writing – original draft, Writing – review & editing, Supervision, Project administration, Funding acquisition.

**Declaration of Competing Interest**
The authors declare that there is no conflict of interest.

**Data Availability**
The data will be made available on reasonable request to the corresponding author.